# Software Requirement Specification Using Reverse Speech Technology

Santhy Viswam[#1], Sajeer Karattil[#2]

[#1] *M.tech student (CSE), MES College of Engineering, Kuttippuram, India*

[#2]*Assistant Professor (CSE), MES College of Engineering, Kuttippuram, India*

*Abstract*— Speech analysis had been taken to a new level with the discovery of Reverse Speech (RS). RS is the discovery of hidden messages, referred as reversals, in normal speech. Works are in progress for exploiting the relevance of RS in different real world applications such as investigation, medical field etc. In this paper we represent an innovative method for preparing a reliable Software Requirement Specification (SRS) document with the help of reverse speech. As SRS act as the backbone for the successful completion of any project, a reliable method is needed to overcome the inconsistencies. Using RS such a reliable method for SRS documentation was developed.

*Keywords*— Reverse Speech, Software Requirement Specification (SRS), Speech Enhancement, Speech Recognition.

## I. INTRODUCTION

Speech is an integral part of human communication. Speech synthesis and analysis and its recognition have been studied for years and still works are going on for the development of a system which is 100% reliable. Speech analysis has been taken to a new level with the discovery of the new technology- Reverse Speech. Reverse Speech is the phenomenon of hidden backward messages in speech. If human speech is recorded and played backwards, mixed amongst the gibberish at regular intervals very clear statements can be heard. These statements usually appear in short sentence form and are nearly always related to the forward speech. Thus by analysing ones' speech we could find the innermost thought of the person.

Automatic speech recognition (ASR) is yet another speech analysis method. ASR can be defined as the independent, computer-driven transcription of spoken language into readable text in real time. In a nutshell, ASR is technology that allows a computer to identify the words that a person speaks into a microphone or telephone and convert it to written text, with 100% accuracy, all words that are intelligibly spoken by any person, independent of vocabulary size, noise, speaker characteristics or accent.

Current systems and methods in real world applications are never 100% reliable which demands for developing better methods and techniques to increase the reliability. Works are progressing in the above mentioned speech analysis techniques that can bring forward many reliable systems in real world such as in the field of investigation, medicine etc. This makes way for a novel approach exploiting the applicability of reverse speech in the software development.

## II. REVERSE SPEECH

A radical new theory about language is enlightened by the discovery of a new way of communication, the Reverse Speech [1]. David John Oates, the founder of Reverse Speech, claimed that embedded backwards into the sounds of human speech there is another form of human communication. The new way of communication says that as the human brain is constructing the sounds of speech, it is forming those sounds in such a way that occasionally another message is said backwards. This second message can be heard if human speech is recorded and played backwards. Mixed among the normal gibberish of backwards audio at regular intervals can be heard quite clear grammatically correct sentences. Furthermore these backward sentences often have a unique relationship to what is being spoken forwards at the time. They can confirm the forward speech, or they can be contradictory to it. Sometimes they can expand upon the forward speech, giving extra information or clarifying details. They may even reveal the thoughts of the speaker at the time.

Typically most of the backward messages found have a direct relationship to both the forward dialogue and the situation at hand. This is called the principal of Speech Complementarity. The theory that describes both the principal of Reverse Speech and its Speech Complementarity [1], [2] goes like this:

1. Human speech has two separate, yet complementary, functions and modes. One mode occurs overtly, is spoken forwards, and is under conscious control. The other mode occurs covertly, is spoken backwards and is not under conscious control. The backward mode of speech occurs simultaneously with the forward mode and is a reversal of the forward speech sounds.
2. These two modes of speech, forward and backward, complement and are dependent upon each other. One mode cannot be fully understood without the other mode. In the dynamics of interpersonal communication both modes of speech combined communicate the total psyche of the person, conscious and unconscious.

3. Covert speech develops before overt speech and children speak backwards before they do forwards. Then, as forward speech begins, the two modes of speech gradually combine into one forming an overall bi-level communication process.

The frequency of reversals that can occur is directly dependent upon the type of conversation or speech that is happening in the forward dialogue. In casual, relaxed conversations speech reversals usually occur once in every 15-30seconds. This can increase, if the rapport between the persons speaking is high, to as frequently as once every 3-5 seconds while speech reversals are rare in public scripted speeches, occurring sometimes as little as once every ten minutes or so. This observation was found consistent across the analysis of literally 1000s of hours of tape, ranging from one on one session work, to private conversations, to telephone recordings and public speeches.

*A. The Layers of Consciousness*

Reverse Speech comes from many different parts of the psyche. It is not just one part of the mind speaking. At times reversals can reflect conscious thoughts that the speaker is aware of having at the time, or emotions they were feeling. At other times they can communicate deeper thoughts of which the speaker has no conscious awareness. Based on the many years of work on Reverse Speech, the human psyche was divided into five distinct parts by David John [1].

1) *Part One- The Conscious Mind:* This is the part of mind that is in the area of consciousness.
2) *Part Two- The Subconscious Mind:* This is the area of the mind that is just below the level of consciousness. It contains memories that were once conscious but have become repressed due to the passage of time or psychological factors.
3) *Part Three- The Deep Unconscious Mind:* There is a level of the mind that is deeper than the surface subconscious mind. This level contains the codes and building blocks of behaviour and personality. It communicates these codes, or patterns, using metaphors.
4) *Part Four- The Collective Unconscious:* This collective unconscious can be seen on rare occasions in Reverse Speech when people show knowledge of events outside their conscious awareness, past, present, and even future events in their speech reversals.
5) *Part Five- The Spirit:* There are many reversals in Reverse Speech that seem to come from a deeper part of us that is guiding and helping, giving advice and instructions. Sometimes it is called the small still voice within.

*B. The Different Types of Reversals*

Most proportion of the reversals is related to the forward dialogue, that is, they are contemporary reversals. The exemption seems to be when communicating collective unconscious information. Exact categorization of the reversals [1], [4] gives them a greater understanding of meaning in the context. The categories are listed below:

- *Congruent reversals*
  These reversals are congruent with and confirm the content of the forward dialogue.
- *Incongruent reversals*
  These reversals will directly contradict what has been said forwards.
- *Expansive reversals*
  These reversals add additional information to the forward dialogue.
- *External dialogue*
  These reversals will directly speak to another person giving instructions or asking questions. It is believed that the unconscious hears these reversed messages, recognizes it as intuition and responds accordingly.
- *Internal dialogue*
  These reversals will reflect internal thought processes, conscious and unconscious, as the psyche talks to themselves, organizing behaviour and self-analysis.
- *Internal Command*
  These reversals will give a direct command concerning some action that must be taken by the conscious self. It is assumed that these commands are coming from some deeper part of the psyche.
- *Trail and Lead reversals*
  These reversals will appear in a conversation either before or after a topic is discussed in forward speech.
- *Future tense reversals*
  These reversals predict a future event or behavioural outcome in a person's life. In the many years of doing Reverse Speech, David John has found them to be accurate.
- *Comparative reversals*
  These reversals have no obvious connection with the subject matter of the forward dialogue, but will rather contain an emotional comparison. That is, the reversal may discuss an event or subject that, although unrelated to the subject of the forward dialogue, has similar emotions attached to it.
- *Constant reversals*
  There are certain words and phrases in Reverse Speech that frequently reverse to say the same thing. For example, America will frequently reverse to say *the crime*.
- *Layered reversals*
  There are a small number of reversals that seem to defy explanation. This is when two different things can be heard on the same section of tape depending on what you are listening for at the time. By shifting your listening perspective, one reversal can be heard, and then another. This is similar to some types of holograms.

Reverse speech has been widely used in forensic applications, business and corporate applications, medical applications like psychotherapy, politics etc.

### III. SPEECH ENHANCEMENT TECHNIQUES

Speech enhancement techniques aim at improving the quality and intelligibility of speech that has been degraded by noise [5], [6], [14]. The goal of speech enhancement varies according to the needs of specific applications, such as to increase the overall speech quality or intelligibility, to reduce listener fatigue or to improve the global performance of an ASR embedded in a voice communication system. So the noisy speech signal is preprocessed by a speech enhancement algorithm before being fed to the speech recognizer. Numerous techniques have been proposed in the literature for speech enhancement. These techniques can roughly be divided into four [7] main categories: spectral subtractive, statistical-model-based, subspace decomposition and perceptual based techniques.

#### A. Spectral Subtractive Techniques

One of the most popular methods of reducing the effect of background (additive) noise is Spectral Subtraction. Spectral subtraction simply needs an estimate of the noise spectrum during periods of speaker silence (single channel) or from a reference source (multichannel). It is a frame-based approach that estimates the short-term spectral magnitude of the noise-free signal from the noisy data. In spectral subtraction, the average noise spectrum is subtracted from the average signal spectrum, performed independently in the frequency bands critical to human hearing. This reduces the level of each frequency band by an amount proportional to the level of that frequency in the noise sample. Suppose the speech signal $x(m)$ is corrupted by background noise $n(m)$ that is:

$$y(m) = x(m) + n(m)$$

Windowing the signal:

$$y_w(m) = x_w(m) + n(m)$$

Fourier transform of both sides

$$Y_w(e^{jw}) = X_w(e^{jw}) + N_w(e^{jw})$$

Where $Y_w(e^{jw})$, $X_w(e^{jw})$ and $N_w(e^{jw})$ are the Fourier transforms of windowed noisy speech and noise signals respectively. The main drawback of these methods is the introduction of an artificial noise called residual noise.

#### B. Statistical Model-based Techniques

Speech enhancement can be approached as a statistical estimation problem. The goal here is to find a linear (or non-linear) estimator of the original clean signal. The Wiener and minimum mean-square error (MMSE) algorithms are among the well-known methods belonging to this category. Wiener filters are considered as linear estimators of the clean speech signal spectrum and they are optimal in the mean-square sense. The enhanced time domain signal is obtained by convolving the noisy signal with a linear (Wiener) filter. Equivalently, in the frequency domain, the enhanced spectrum is obtained by multiplying the input noisy spectrum by the Wiener filter.

#### C. Subspace Decomposition Techniques

These techniques are based on the principle that a nonparametric linear estimate of the unknown clean-speech signal is obtained by using a decomposition of the observed noisy signal into mutually orthogonal signal and noise subspaces. This decomposition is performed under the assumption that the energy of less correlated noise spreads over the entire observation space while the energy of the correlated speech components is concentrated in a subspace generated by the low-order components. The noise is assumed to be additive and uncorrelated with speech signal. Noise reduction is obtained by removing the noise subspace and by removing the noise contribution in the signal subspace. The decomposition of the vector space of the noisy signal into subspaces can be done using the well-known orthogonal matrix factorization techniques from linear algebra namely, the Singular Value Decomposition (SVD) or the Eigen Value Decomposition (EVD).

#### D. Perceptual-based Techniques

The goal of perceptual-based methods is to make the residual noise perceptually inaudible and therefore to improve the intelligibility of enhanced signals by considering the properties of the human auditory system. The idea is to exploit the fact that the hearing system cannot perceive residual noise when its level falls below the noise masking threshold (NMT). In these methods, the spectral estimates of a speech signal play a crucial role in determining the value of the noise masking threshold that is used to adapt the perceptual gain factor.

### IV. MFCC FEATURE EXTRACTION TECHNIQUE

Cepstral based features are the most commonly used feature vectors in speech processing tasks. The representation of a speech signal in the cepstral domain helps us to model the speech signal as source filter model. The most widely used feature vector based on cepstrum is MFCC. The advantage of MFCC is that it takes into account the perceptual characteristics of the human ear. The frequencies perceived by the human ear are in a nonlinear logarithmic scale rather than in a linear scale and the frequencies are perceived in a nonlinear frequency binning (critical band filtering). The nonlinear scale is characterized by Mel scale and critical band filtering is characterized by Mel filter bank. The MFCC feature extraction technique [8], [9] basically includes windowing the signal, applying the DFT, taking the log of the magnitude and then warping the frequencies on a Mel scale, followed by applying the inverse DCT. Various steps involved in the MFCC feature extraction is listed below:

1) *Pre-emphasis:*
Pre-emphasis refers to filtering process designed to increase the magnitude of usually higher frequencies with respect to the magnitude of lower frequencies in order to improve the overall signal-to-noise ratio. Commonly used pre-emphasis filter is given by the following transfer function.

$$H(z) = 1 - a^* z^{-1}$$

The goal of pre-emphasis is to compensate the high-frequency part that was suppressed during the sound production.

2) *Frame Blocking:*
In order to make speech signal stationary, we split the signal into small blocks so that the signal is stationary in that segment. It means that when speech signal is examined over a short period of time it has quite stable acoustic characteristics. The input speech signal is segmented into frames of 20-30 ms with optional overlap of 1/3-1/2 of the frame size.

3) *Hamming Window:*
We multiply each frame with window function in order to keep the continuity of the first and the last points in the frame. The most commonly used windowing function is hamming and hanning window. If the signal in a frame is denoted by $s(n), n = 0,...,N-1$, then the signal after Hamming windowing is $s(n)*w(n)$, where $w(n)$ is the Hamming window defined by:

$$w(n,a) = (1-a) - a\cos(2\pi n/(N-1)), 0 \le n \le N-1$$

Windowing is performed in order to enhance the harmonics, smooth the edges and to reduce the edge effect while taking the DFT on the signal.

4) *DFT Spectrum:*
Each windowed frame is converted into magnitude spectrum by applying,

$$X(k) = \sum_{n=0}^{N-1} x(n) e^{\frac{-j2\pi nk}{n}} ; \quad 0 \le k \le N-1$$

where N is the number of points used to compute the DFT.

5) *Mel-frequency wrapping:*
Human perception of frequency contents of sounds for speech signal does not follow a linear scale. Thus for each tone with an actual frequency, $f$, is measured on a scale called the 'mel' scale. The mel-frequency scale is linear frequency spacing below 1000 Hz and a logarithmic spacing above 1000Hz. The approximation of mel from physical frequency can be expressed as:

$$f_{mel} = 2595 \log_{10}\left(1 + \frac{f}{700}\right)$$

where $f$ is the physical frequency and $f_{mel}$ denotes the perceived frequency.

6) *Discrete Cosine Transform:*
The DCT is applied to the transformed mel-frequency coefficients produces a set of cepstral coefficients. Since most of the signal information is represented by the first few MFCC coefficients, the system can be made robust by extracting only those coefficients ignoring or truncating higher order DCT components. Finally, MFCC is calculated as:

$$c(n) = \sum_{m=0}^{M-1} \log_{10}(s(m)) \cos\left(\frac{\pi n(m-0.5)}{M}\right)$$

for $n = 0,1,2,...,C-1$ where $c(n)$, the cepstral coefficients and $C$, the number of MFCCs.

7) *Dynamic MFCC Features:*
The time derivatives of cepstral coefficients, the first and second derivatives, are known as velocity and acceleration features. The first order derivative is also called delta coefficients that tell us about the speech rate, and the second order derivative is called delta-delta coefficients that give information similar to acceleration of speech. The dynamic parameter can be calculated as:

$$\Delta c_m(n) = \frac{\sum_{i=-T}^{T} k_i c_m(n+i)}{\sum_{i=-T}^{T} |i|}$$

V. GAUSSIAN MIXTURE MODEL

In the speech and speaker recognition the acoustic events are usually modelled by Gaussian probability density functions (PDFs) [9], described by the mean vector and the covariance matrix. However uni-model PDF with only one mean and covariance are unsuitable to model all variations of a single event in speech signals. Therefore, a mixture of single densities is used to model the complex structure of the density probability.

A Gaussian mixture model is a weighted sum of M component Gaussian densities as given by the equation,

$$p(x|\lambda) = \sum_{i=1}^{M} w_i g(x|\mu_i, \Sigma_i),$$

where $x$ is a D-dimensional continuous-valued data vector, features, $w_i$, $i=1,...,M$ are the mixture weights, and $g(x|\mu_i, \Sigma_i)$, $i=1,...,M$, are the component Gaussian densities. Each component density is a D-variate Gaussian function of the form,

$$g(x|\mu_i, \Sigma_i) = \frac{1}{(2\pi)^{D/2} |\Sigma_i|^{1/2}} \exp\left\{-\frac{1}{2}(x-\mu_i)' \Sigma_i^{-1} (x-\mu_i)\right\}$$

with mean vector $\mu_i$ and covariance matrix $\Sigma_i$. The mixture weights satisfy the constraint that $\lambda = \{w_i, \mu_i, \Sigma_i\}$, $i=1,...,M$.

## VI. PROPOSED WORK

In the Software Development Life Cycle (SDLC) requirement acquisition phase forms the backbone for the successful completion of the project. SRS documentation encounters many problems that lead to the failure of the project at the time of deployment. A novel and judicious approach to deal with the problem is made possible using Reverse Speech technology.

*A. Applying Reverse Speech*

The Reverse Speech technology removes the inconsistencies in the forward speech using the backward speech. As mentioned earlier it is the phenomenon of hidden backward messages in speech. So by introducing reverse speech for SRS preparation we are able to document SRS very precise, clear, correct and genuine. More the emotional content, clearer is the reversals obtained from the conversation with the customer. Thus obtained reversals make the requirements clear. Moreover the additive noise that may occur due to electronic devices and the environment is removed using by filtering and the speech enhancement results in clean speech signals.

*B. Recognition*

The recognition module in the work recognizes the forward speech which is recorded and saved in .wav file format. The file is passed with the enhancement module for removal of the additive noise in the recorded file. The reversals obtained as output from the application of the Reverse Speech is also enhanced by additive noise removal technique. Both the forward and reverse speech obtained are recognized and documented that clearly reveal the inconsistencies in the recorded forward speech with the help of the recognized reverse speech.

## VII. EXPERIMENTAL RESULTS

On working with different data and reversing the record there were some reversals obtained and these were analyzed by different people to ensure the reversals obtained. The ReverseSpeech Pro 2.6 version and Matlab coding were used to get the reversals. Examples of reversals thus obtained are listed below:

Forward: 'Everybody yeah....'(part of a song)
Reverse: 'Thy hate on me why'

Forward: 'I am happy'(said by a person)
Reverse: 'We can't fail'.

These reversals are taken as the input to the speech recognition module and the speech recognition module extracts the features of the speech using MFCC and thus obtained features are used for the recognition of the speech. And thus the speech to text conversion takes place. By using the MFCC the feature extraction is made easier as it produces a 39-dimension matrix that can be considered as different features extracted.

## VIII. CONCLUSIONS

The requirement acquisition and analysis phase needs to be robust and reliable. In current situation about 25% of the software fails due to lack of proper understanding of the actual requirement of the customer. The proposed work uses the reverse speech technology as an efficient and reliable system for SRS document preparation. The reverse speech technology acts like a modality healing. The work outputs the SRS documentation after recognizing the forward as well as the backward speech. And with the help of the documentation of the forward and reverse, the inconsistencies in the gathered requirement, from the conversation, and actual requirement of the user or customer are removed.

ACKNOWLEDGMENT

I wish to thank Sajeer Karattil, Dr. Raghav Menon, Dr. Paul Rodingous and all helping hands that made this work a successful one.